\documentclass[lettersize,journal]{IEEEtran}
\usepackage{amsmath,amsfonts}
\usepackage{algorithmic}
\usepackage{algorithm}
\usepackage{array}
\usepackage[caption=false,font=normalsize,labelfont=sf,textfont=sf]{subfig}
\usepackage{textcomp}
\usepackage{stfloats}

\usepackage{url}
\usepackage{verbatim}
\usepackage{graphicx}
\usepackage{gensymb}
\usepackage{amssymb}
\usepackage{esvect}
\usepackage{soul}
\usepackage{color}
\usepackage{multirow} 
\usepackage{diagbox} 
\usepackage{pdfpages}
\usepackage[numbers,sort&compress]{natbib}
\usepackage[colorlinks,linkcolor=blue,anchorcolor=blue,citecolor=blue]{hyperref}

\usepackage[colorlinks,linkcolor=blue,anchorcolor=blue,citecolor=blue]{hyperref}
\begin{document}
\bstctlcite{IEEEexample:BSTcontrol}

\title{\LARGE \bf Inductance-Based Force Self-Sensing in Fiber-Reinforced Pneumatic Twisted-and-Coiled Actuators}
%Exploiting Low-Hysteretic Inductance-Force Mapping for Self-Sensing of FR-PTCA"
%Exploiting Inherent Low-Hysteresis Inductance-Force Coupling for Self-Sensing in FR-PTCA
%Inductance-Based Force Self-Sensing via Intrinsic Low-Hysteresis Electromechanical Coupling in FR-PTCAs
%Force Self-Sensing in FR-PTCA via Intrinsic Low-Hysteretic Inductance-Force Coupling
%Inductance as a Force Predictor: Low-Hysteretic Self-Sensing for Fiber-Reinforced Pneumatic Twisted and Coiled Actuators
%Simultaneous Force and Displacement Estimation for FR-PTCA via Inductance-Based Self-Sensing with Hysteresis Mitigation

\author{Yunsong Zhang, Tianlin Li, Mingyang Yang and Feitian Zhang\textsc{*}
%,~\IEEEmembership{Member,~IEEE,}
        % <-this % stops a space
%\thanks{\hl{This paper was produced by the IEEE Publication Technology Group. They are in Piscataway, NJ.}}% <-this % stops a space
\thanks{The authors are with the Robotics and Control Laboratory, School of Advanced Manufacturing and Robotics, and the State Key Labora tory of Turbulence and Complex Systems, Peking University, Beijing 100871, China (e-mail: zhangyunsong@stu.pku.edu.cn; ltl2897087018@stu.pku.edu.cn; mingyangyang@stu.pku.edu.cn; feitian@pku.edu.cn).}
%\thanks{Yunsong Zhang is with the Department of Advanced Manufacturing and Robotics, College of Engineering, Peking University, Beijing, 100871, China,{\tt\small zhangyunsong@stu.pku.edu.cn}}
%\thanks{Tianlin Li is with the Department of Advanced Manufacturing and Robotics, College of Engineering, Peking University, Beijing, 100871, China,{\tt\small zhangyunsong@stu.pku.edu.cn}}
%\thanks{Mingyang Yang is with the Department of Advanced Manufacturing and Robotics, College of Engineering, Peking University, Beijing, 100871, China,{\tt\small zhangyunsong@stu.pku.edu.cn}}
%\thanks{Feitian Zhang is with the Department of Advanced Manufacturing and Robotics, and the State Key Laboratory of Turbulence and Complex Systems, College of Engineering, Peking University, Beijing, 100871, China, {\tt\small feitian@pku.edu.cn}}
\thanks{* Send all correspondences to F. Zhang.}}

% The paper headers
\markboth{}%
{Shell \MakeLowercase{\textit{et al.}}: A Sample Article Using IEEEtran.cls for IEEE Journals}

\maketitle
\begin{abstract}

Fiber-reinforced pneumatic twisted-and-coiled actuators (FR-PTCAs) offer high power density and compliance but their strong hysteresis and lack of intrinsic proprioception limit effective closed-loop control. This paper presents a self-sensing FR-PTCA integrated with a conductive nickel wire that enables intrinsic force estimation and indirect displacement inference via inductance feedback. Experimental characterization reveals that the inductance of the actuator exhibits a deterministic, low-hysteresis inductance-force relationship at constant pressures, in contrast to the strongly hysteretic inductance-length behavior. Leveraging this property, this paper develops a parametric self-sensing model and a nonlinear hybrid observer that integrates an Extended Kalman Filter (EKF) with constrained optimization to resolve the ambiguity in the inductance-force mapping and estimate actuator states. Experimental results demonstrate that the proposed approach achieves force estimation accuracy comparable to that of external load cells and maintains robust performance under varying load conditions.

\end{abstract}


\section{Introduction}

\IEEEPARstart{A}{rtificial} muscles have attracted significant attention in the field of robotics due to their inherent compliance and high power-to-weight ratios \cite{zhang2019robotic}. Among various actuation technologies, Twisted and Coiled Actuators (TCAs) have emerged as a prominent class owing to their high work density and low manufacturing cost \cite{haines2014artificial, zhang2019robotic, zhang2021design}. However, conventional TCAs are typically driven by electrothermal actuation, which leads to limited energy efficiency and slow dynamic responses \cite{haines2014artificial, bao2023fast}. 

To address these limitations, fluid-driven variants, such as the Cavatappi artificial muscle, have been proposed \cite{higueras2021cavatappi,higueras2022material}. More recently, Weissman \textit{et al.} introduced the Fiber-Reinforced Pneumatic TCA (FR-PTCA), which employs fiber reinforcement to enhance structural anisotropy \cite{weissman2025efficient}. This design significantly improves energy efficiency (exceeding $19\%$) and achieves up to $70\%$ contraction strain at relatively low pressures, making the FR-PTCA a promising actuation solution for high-performance soft robotic systems.

Despite these mechanical advances, the practical deployment of FR-PTCAs remains limited by the lack of intuitive modeling and sensing frameworks for real-time control \cite{higueras2022material, weissman2025efficient}. Unlike thermal TCAs, for which linear time-invariant models have been derived to facilitate feedback control \cite{yip2017control, sun2023development}, pneumatic TCAs currently lack a unified modeling approach. This challenge becomes particularly pronounced in soft continuum robots, where accurate state estimation is difficult. Conventional rigid sensors, such as encoders or external motion capture systems, introduce additional mass and mechanical stiffness, which compromises the compliance adaptability of soft robotic systems \cite{rus2015design}. 
Consequently, the development of FR-PTCAs with integrated self-sensing capabilities, whereby the actuator simultaneously functions as a sensor, is critical. Such capability would enable accurate state estimation and closed-loop control while preserving the fundamental compliance of robotic systems \cite{huang2025precise, sun2023development, su2024spatial}.

\begin{figure}[htbp]
        \centering
        \includegraphics[width=0.5\textwidth]{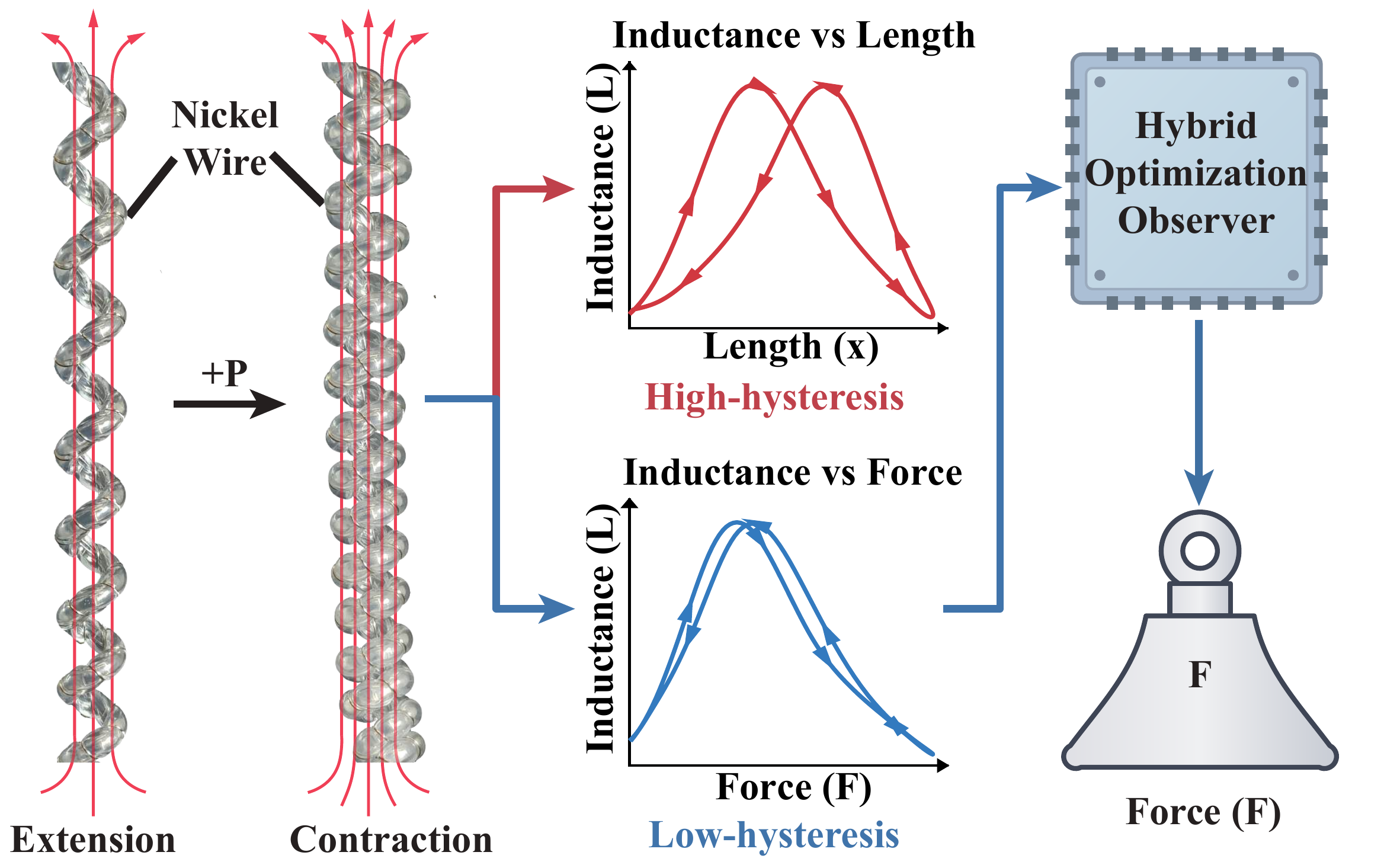}
        \caption{Conceptual overview of the proposed inductive self-sensing framework for FR-PTCAs. A helically wrapped conductive nickel wire enables intrinsic inductive sensing without external rigid sensors. While the inductance-length relationship exhibits strong hysteresis, the inductance-force mapping remains deterministic and low-hysteresis at constant pressures. Leveraging this property, a nonlinear hybrid observer processes inductance measurements to enable real-time estimation of actuator force and internal states.}
        \label{Fig1}
\end{figure}

%Self-sensing refers to the capability of extracting actuator state information by exploiting its intrinsic electrical or mechanical properties \cite{dosch1992self, kruusamae2015self}. This concept has been widely investigated in artificial muscle systems. For example, piezoelectric actuators utilize capacitance for position estimation \cite{dosch1992self}, while shape memory alloys and thermal TCAs typically rely on electrical resistance to infer temperature and strain \cite{ikuta1988shape, ma2004position, tang2019temperature, bombara2021experimental, huang2025precise}. Regarding pneumatic artificial muscles (PAMs), several approaches have been explored, including inductive sensing using embedded springs \cite{cho2023bidirectional}, capacitance-based sensing via dielectric elastomer sensors \cite{kanno2021self}, and magnetic field-based sensing for displacement and force estimation \cite{su2025embedded}. In the context of thermal TCAs, electrical impedance has also been employed to estimate displacement and temperature \cite{van2016self}. However, these approaches often suffer from strong thermal coupling, temperature drift, and hysteresis, which complicate accurate state estimation \cite{van2016self, van2019closed, sun2020integrated}. In contrast, fluid-driven FR-PTCAs avoid these thermal complications, providing an opportunity to exploit inductive sensing for stable, high-fidelity self-sensing. 

Self-sensing refers to the capability of extracting actuator state information by exploiting its intrinsic electrical or mechanical properties \cite{dosch1992self}. This concept has been widely investigated in artificial muscle systems. For example, piezoelectric actuators utilize capacitance for position estimation \cite{dosch1992self}. Thermal TCAs typically rely on electrical resistance or impedance to infer temperature and strain \cite{tang2019temperature, bombara2021experimental, huang2025precise, van2016self}. However, these approaches often suffer from strong thermal coupling, temperature drift, and hysteresis, which complicate accurate state estimation \cite{van2016self, van2019closed, sun2020integrated}. Regarding pneumatic artificial muscles (PAMs), several approaches have been explored, including inductive sensing using embedded springs \cite{cho2023bidirectional}, capacitance-based sensing via dielectric elastomer sensors \cite{kanno2021self}, and magnetic field-based sensing for displacement and force estimation \cite{su2025embedded}. Sharing the twisted helical geometry of thermal TCAs and the fluid-driven actuation of PAMs, FR-PTCAs represent a unique technological intersection. By inherently bypassing the thermal complications present in traditional TCAs, FR-PTCAs offer a distinct opportunity to exploit inductive sensing for stable, high-fidelity self-sensing.

Building upon the structural assumptions proposed by Weissman \textit{et al.} \cite{weissman2025efficient}, namely the inextensibility of the reinforcement fiber and the no-slip condition between the fiber and the tube, the FR-PTCA can be viewed as a nested double-helix structure comprising a macroscopic helical coil and a microscopic twisted fiber trajectory. Under these kinematic constraints, we hypothesize that the integration of conductive reinforcing fibers maintains tight coupling with actuator deformation. As illustrated in Fig.~\ref{Fig1}, this configuration enables the actuator inductance to couple preferentially with internal tension and output force rather than with the macroscopic actuator length. Experimental observations suggest that actuator inductance exhibits a deterministic and low-hysteresis relationship with the output force at constant pressures, in contrast to the strongly hysteretic inductance-length relationship. This physical property provides the basis for a reliable force self-sensing mechanism and motivates the observer design developed in this work.

This paper presents a self-sensing FR-PTCA and a unified framework for its fabrication, modeling and control. We replace the conventional nylon monofilament with a conductive nickel wire, endowing the actuator with intrinsic inductive sensing capability while preserving its mechanical compliance. Experimental characterization reveals a deterministic, low-hysteresis relationship between actuator inductance and output force at constant pressures. A dynamic model is established to characterize the interplay between force, pressure, and length. Leveraging this property, we develop a parametric self-sensing model and a nonlinear hybrid observer that integrates an EKF with constrained optimization to estimate actuator states. Finally, inductance-based feedback control is implemented and experimentally evaluated against open-loop and sensor-based control strategies. The results demonstrate that the proposed self-sensing method achieves force estimation accuracy comparable to external load cells and significantly improves tracking accuracy under varying load conditions.

The main contributions of this paper are threefold. First, the design and fabrication of a self-sensing FR-PTCA are presented, providing intrinsic proprioception while preserving mechanical compliance. Second, dynamic and self-sensing models are established to characterize the low-hysteresis coupling between the inductance, internal pressure, and output force. Third, a hybrid EKF-optimization observer is developed to facilitate real-time state estimation by resolving non-monotonicity in the inductance mapping. This integrated framework provides a robust solution for the control of soft robotic systems without the need for external instrumentation.

\section{Design and Fabrication}

\begin{figure}[htbp]
        \centering
        \includegraphics[width=0.45\textwidth]{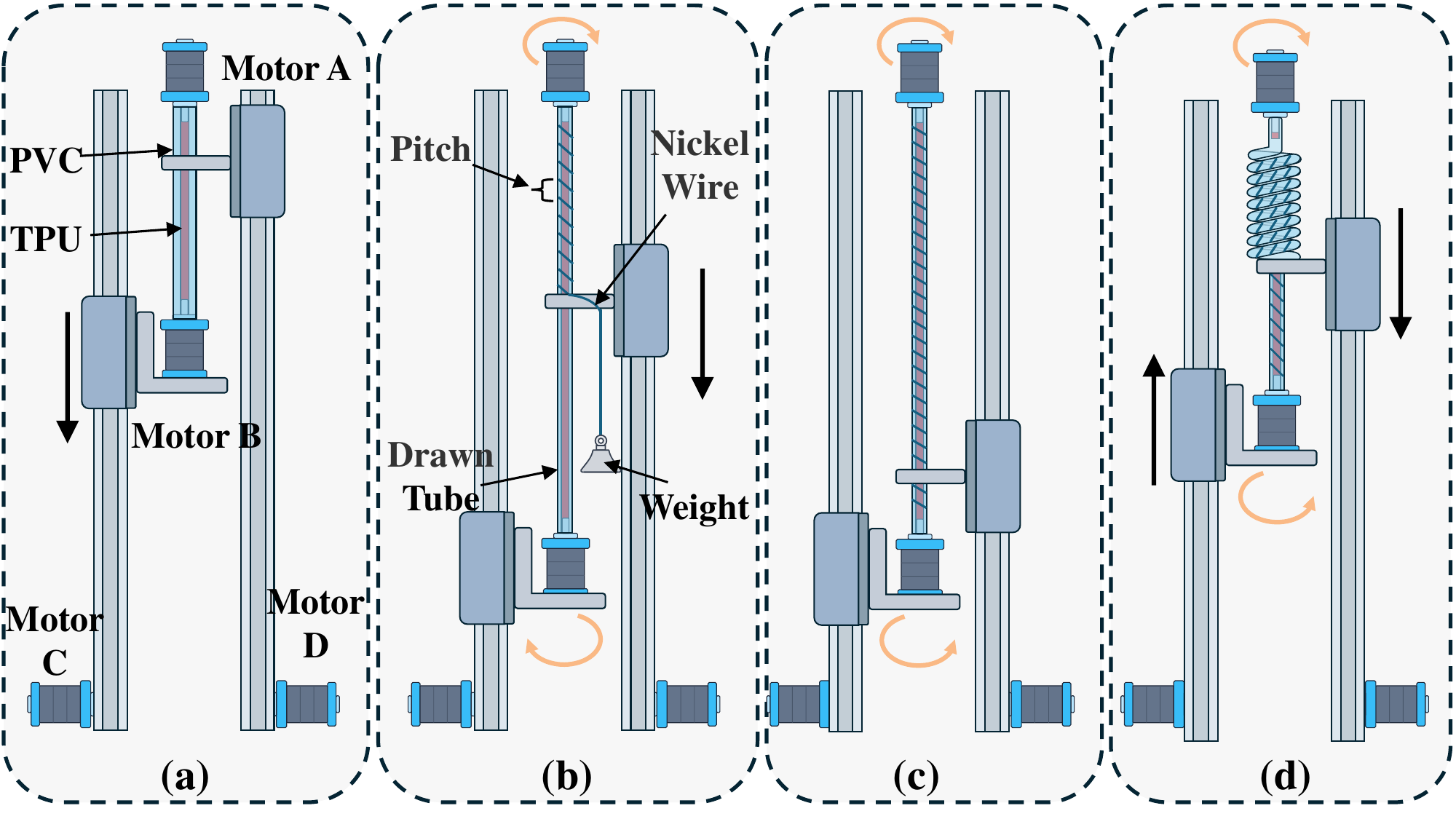}
        \caption{Fabrication process of the self-sensing FR-PTCA. Schematics of the fabrication platform and its stages: (a) precursor preparation and drawing, (b) conductive fiber wrapping, (c) twisting, (d) coiling. }
        \label{Fig2}
\end{figure}

To enable inductance-based self-sensing, the fabrication protocol of the standard FR-PTCA \cite{weissman2025efficient} was modified by replacing the conventional reinforcing fiber with a conductive nickel wire. The fabrication process consists of six stages: 1) precursor preparation, 2) drawing, 3) conductive fiber reinforcement, 4) twisting, 5) coiling, and 6) annealing. As illustrated in the schematic of the custom fabrication platform in Fig. \ref{Fig2}, the apparatus incorporates two rotary stepper motors (Motors A and B) to apply torque, along with two linear stepper motors equipped with guide rails (Motors C and D) to control axial displacement and tension.

Initially, a thermoplastic polyurethane (TPU) filament (diameter $1.75$ mm, PolyFlex TPU95-HF) was inserted into an undrawn polyvinyl chloride (PVC) tube ($5/32$'' OD $\times$ $3/32$'' ID, ND-100-65 Tygon). Both ends of the PVC tube were secured using pneumatic rapid fittings. Motor D was then used to stretch the PVC tube to a draw ratio of $\lambda = 2.5$.

The primary distinction from the standard fabrication lies in the fiber reinforcement stage. A conductive nickel wire (diameter $0.25$ mm) was helically wrapped around the pre-stretched PVC tube under a constant tension of $100$\,g, selected for its mechanical flexibility and stable electrical conductivity. A winding pitch of $6$ mm was selected to balance a key trade-off: an excessively large pitch weakens the inductance-deformation coupling, whereas an overly small pitch restricts the tube's radial expansion and reduces the achievable contraction stroke.

In the twisting stage, the composite was twisted with the same chirality as the nickel wire to induce a polymer chain angle of $\alpha = 15^\circ$ relative to the tube axis. This configuration ensures robust mechanical contact between the nickel wire and the tube surface, which is critical for establishing stable electromechanical coupling between inductance variations and actuator deformation.

Finally, the twisted structure was coiled onto a mandrel with a radius of $1.5$ mm and a coiled pitch of $6$ mm. The assembly was subsequently annealed at $90~^\circ$C for $45$ minutes to permanently set the helical structure. This fabrication process yields an FR-PTCA with integrated conductive reinforcement that enables intrinsic inductance-based sensing while preserving the actuator's mechanical compliance.

\section{Physical Modeling for Self-Sensing}

\subsection{Force-Length-Pressure Characterization}
The driving mechanism of the FR-PTCA relies on dual anisotropic constraints imposed by high-stiffness reinforcing fibers and cold-drawn polymer chains. Under pneumatic actuation, the radial expansion of the elastomeric tubing is transformed into an unwinding torque \cite{weissman2025efficient}, which the helical geometry subsequently converts into linear contraction. This behavior is mechanically analogous to thermal TCAs, where heat-triggered anisotropic radial expansion of polymer fibers induces microscopic untwisting and macroscopic axial contraction \cite{yip2017control}. Despite the use of distinct energy sources, the underlying kinematic mechanisms are highly similar. This similarity suggests that control-oriented modeling approaches previously developed for thermal TCAs are transferable to the FR-PTCA system.

\begin{figure}[htbp]
        \centering
        \includegraphics[width=0.45\textwidth]{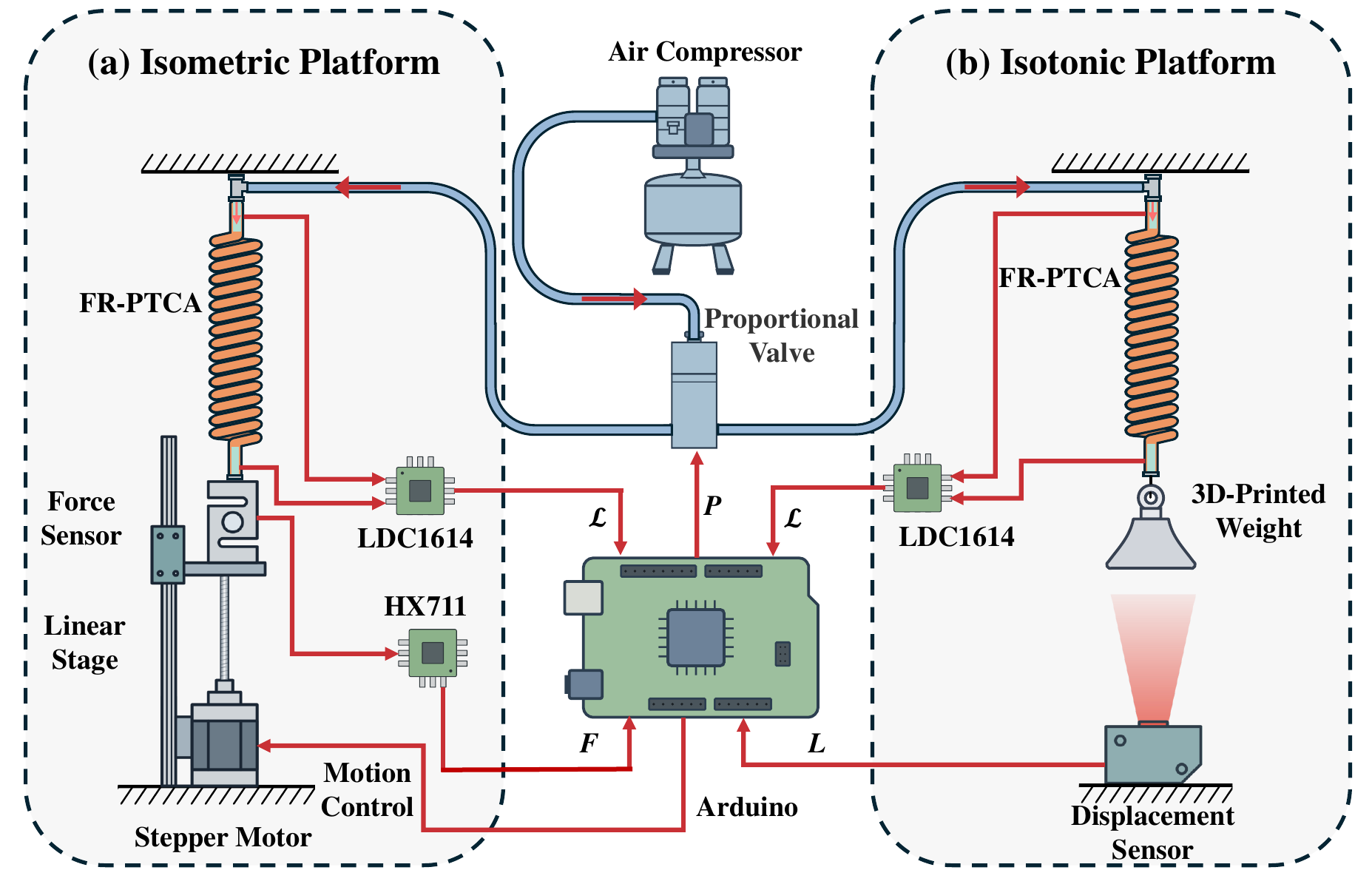}
        \caption{Experimental setup for the FR-PTCA. (a) Isometric Platform. (b) Isotonic Platform.}
        \label{Fig3}
\end{figure}

To characterize the dynamic behavior of FR-PTCA, a custom experimental testbed was developed, as illustrated in Fig.~\ref{Fig3}(a). The setup captures the dynamic relationships among the input pressure $P$, contraction length $x$, and output force $F$. The testbed components include an air compressor (HW51/9, HOSINYON), a proportional valve (ITV1050, SMC), a load cell(AR-DN20, ARIZON), a HX711 signal amplifier, a stepper motor coupled to a linear stage, an inductance measurement chip (LDC1614), and an Arduino micro-controller. The proportional valve precisely regulates the input pressure. Custom 3D-printed connectors secure the FR-PTCA to the load cell while constraining global rotation during inflation, thereby ensuring consistency between the experimental conditions and practical applications. The Arduino unit controls the stepper motor to set the actuator length $x$ while simultaneously acquiring force data and inductance readings from the sensor chip.

\begin{figure}[htbp]
        \centering
        \includegraphics[width=0.5\textwidth]{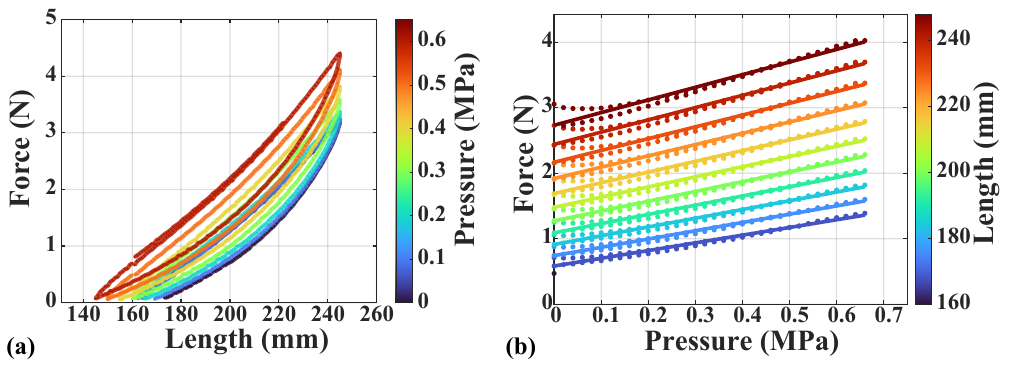}
        \caption{Experimental characterization of the FR-PTCA. (a) Force-length relationship at constant pressures. (b) Force-pressure relationship at fixed lengths.}
        \label{Fig4}
\end{figure}

A series of isometric and isobaric experiments were conducted on FR-PTCAs of varying initial lengths to establish the dynamic model, with the results summarized in Fig.~\ref{Fig4}. Specifically, three samples were tested for each initial length, which ranged from 80 mm to 160 mm in increments of 10 mm. First, isobaric experiments were performed at discrete pressure levels from 0 to 0.65 MPa in increments of 0.05 MPa. At each constant pressure level, the actuator was stretched from a zero-force state to approximately 170\% of the initial length in increments of 2\% for three complete loading and unloading cycles. The force-length response measurements reveal a clear hysteresis loop in the force-strain relationship, as illustrated in Fig.~\ref{Fig4}(a). This behavior aligns with the classical Preisach hysteresis model commonly observed in artificial muscles \cite{tsabedze2021design, yip2017control, zhang2017modeling}. Second, isometric tests were performed at fixed actuator lengths ranging from 100\% to 170\% of the initial length in increments of 5\%. For each fixed length, the internal pressure was cycled five times between 0 and 0.66 MPa in increments of 0.02 MPa while recording the output force. As illustrated in Fig.~\ref{Fig4}(b), the force-pressure relationship is predominantly linear with negligible hysteresis. These observations indicate that the FR-PTCAs exhibit dynamic behavior similar to that of thermal TCAs \cite{yip2017control}. Based on these results, the hysteresis behavior observed in Fig.~\ref{Fig4}(a) is approximated by linearizing the force-length relationship, resulting in the following control-oriented model:

\begin{equation}
F = k(x - x_0) + cP
\label{dyn_model}
\end{equation}
Here, $k$ denotes the stiffness, $x_0$ represents the unloaded actuator length, and $c$ is the pressure coefficient. The parameters were identified via least-squares regression using the experimental data. For a 100 mm actuator, the identified parameters are $k = \text{38.6 N/m}$ and $c = \text{1.6310 N/MPa}$. While this simplified model captures the primary actuator dynamics, it does not provide intrinsic state measurement. Therefore, the next section investigates the electromechanical coupling between inductance and actuator states to enable self-sensing.

\subsection{Inductive Self-Sensing Model}
%Inductance-Force-Pressure Characterization
\begin{figure*}[htbp]
        \centering
        \includegraphics[width=0.9\textwidth]{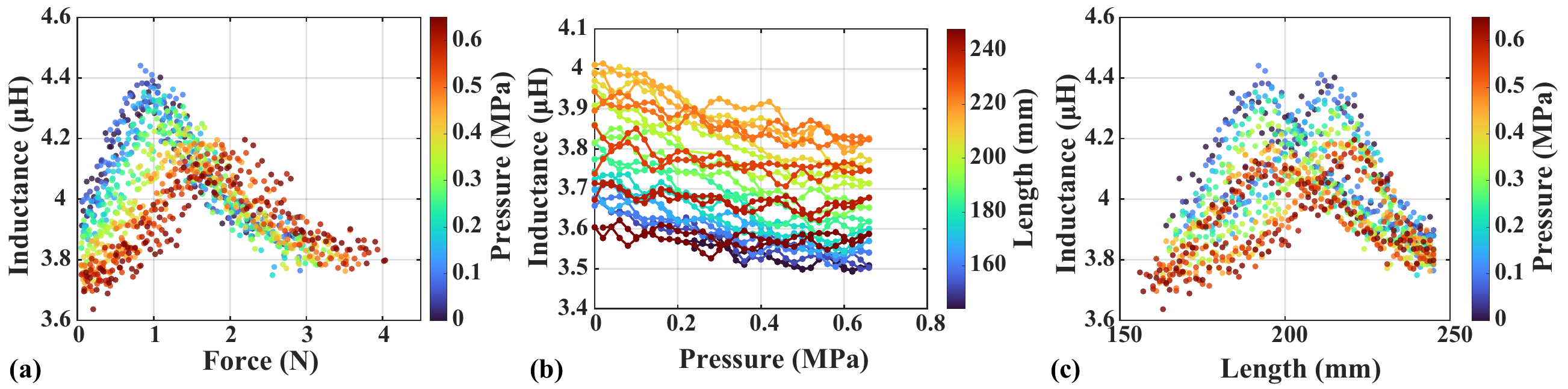}
            \caption{Experimental characterization of the FR-PTCA inductance response. (a) Inductance versus output force at varying internal pressures. (b) Inductance versus internal pressure under isometric conditions for different lengths. (c) Inductance versus actuator length under isobaric conditions.}
        \label{Fig5}
\end{figure*}

The self-sensing principle of FR-PTCA originates from the electromechanical coupling between the conductive reinforcement and the actuator deformation, wherein the geometric reconfiguration of the conductive helix modulates its inductance. From a physical standpoint, the total inductance of the FR-PTCA is governed by the dynamic evolution of both its macroscopic coil structure and the microscopic wire trajectory. As the actuator extends or contracts, variations in the helical radius and pitch induce a electromagnetic field redistribution, resulting in inductance changes that encode the instantaneous mechanical state of the system.

However, experimental characterization reveals behaviors that deviate from conventional linear sensing assumptions \cite{van2016self, van2019closed}. Fig.~\ref{Fig5} illustrates the relationship between inductance and force, pressure, and length. While one might expect a monotonic and deterministic relationship between inductance and actuator length, the experimental results in Fig.~\ref{Fig5}(c) reveal significant hysteresis in the inductance-length relationship under constant pressures conditions. This hysteretic behavior introduces a fundamental ambiguity in state estimation: a single inductance value corresponds to multiple actuator lengths depending on the direction of motion. Resolving this ambiguity requires knowledge of the contraction or extension state; however, paradoxically, identifying the motion direction itself relies on historical length information, thereby creating a circular dependency that renders direct length-based self-sensing unreliable.

In contrast, Fig.~\ref{Fig5}(a) demonstrates that the inductance exhibits a deterministic and low-hysteretic relationship with the output force under constant pressure. This observation suggests that the inductive response of the FR-PTCA is more strongly coupled to the internal mechanical stress state of the conductive reinforcement than to the global axial strain. Similarly, Fig.~\ref{Fig5}(b) indicates a highly deterministic relationship between inductance and pressure under isometric conditions. This distinct characteristic separates the FR-PTCA from thermal TCAs, which typically exhibit strong inductance-length correlations but suffer from substantial force estimation errors due to pronounced strain-force hysteresis \cite{van2016self}. 

Leveraging these observations, we formulate a data-driven self-sensing model that maps inductance to actuator force $F$ and internal pressure $P$. To capture the nonlinear variations observed in the experimental force-inductance curves across different pressures, a parametric model is designed as follows:
\begin{equation}
 \mathcal{L}(F, P) = \lambda_1(P) F^{\lambda_2(P)} e^{\lambda_3(P) F^{\lambda_4(P)}} + \lambda_5(P)
\end{equation}
where $F$ denotes the output force and $\lambda_i(P)$ ($i=1\dots5$) represents pressure-dependent coefficients. These coefficients are modeled as linear functions of internal pressure $P$:
\begin{equation}
\lambda_i(P) = p_{2i-1} P + p_{2i}
\end{equation}
where $p_i$ are identified parameters obtained through experimental calibration. The parameter identification is conducted using the Trust-Region-Reflective least-squares algorithm.

\begin{figure}[htbp]
        \centering
        \includegraphics[width=0.5\textwidth]{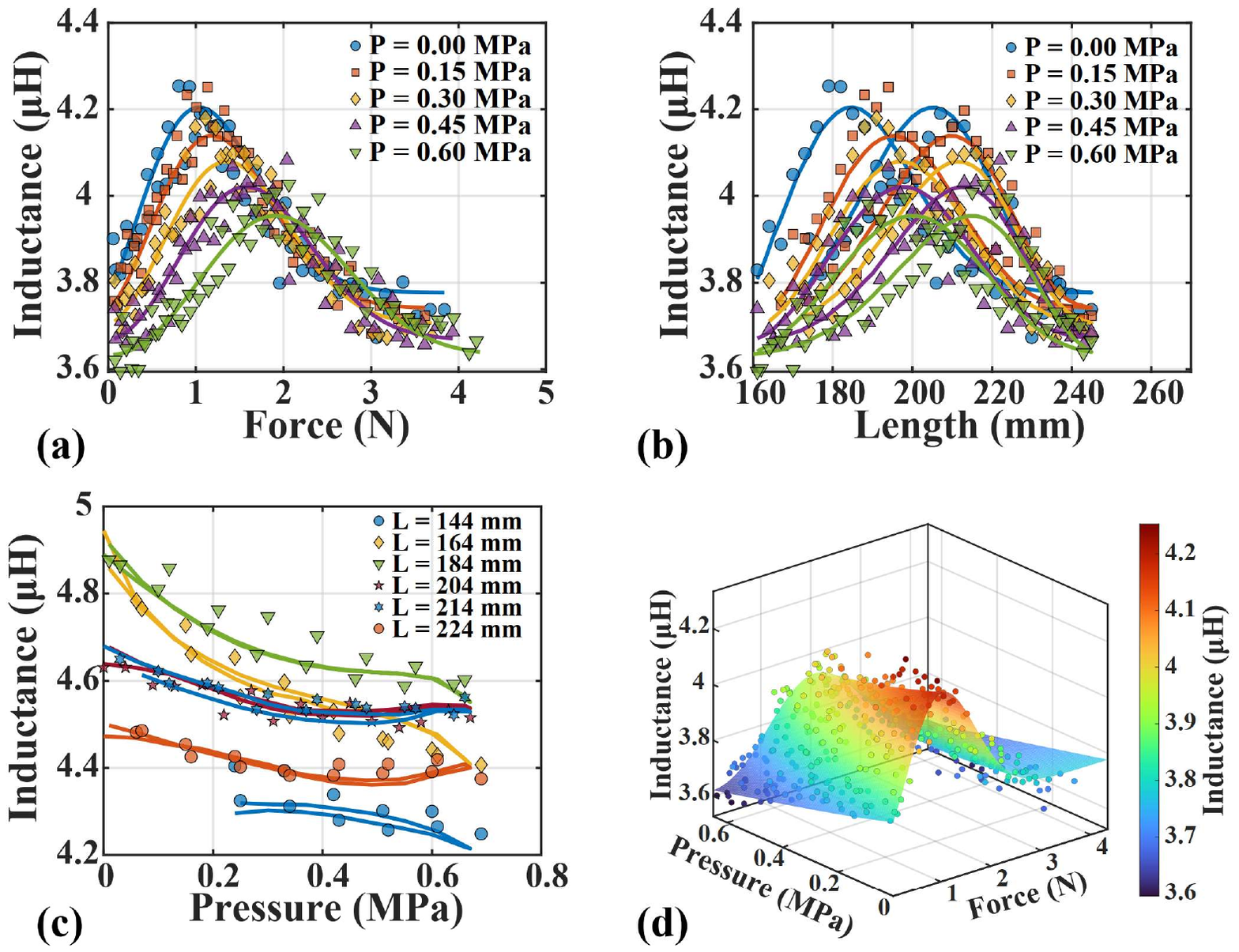}
        \caption{Validation of the proposed data-driven self-sensing model. (a) Experimental versus estimated inductance-force relationship. (b) Model prediction of inductance versus length, illustrating indirect hysteresis compensation. (c) Fitting results of inductance versus pressure. (d) Three-dimensional visualization of the inductance-pressure-force mapping, demonstrating the generalization capability of the model across the workspace.}
        \label{Fig6}
\end{figure}

The performance of the proposed self-sensing model is evaluated using the experimental dataset shown in Fig.~\ref{Fig6}. To verify the generalization of the model, three samples were tested for each initial length, which ranged from 80 mm to 160 mm in increments of 10 mm. For each sample, isobaric stretching experiments were performed at discrete internal pressures from 0 to 0.65 MPa in increments of 0.05 MPa. At each constant pressure level, the actuator was stretched to approximately 170$\%$ of the initial length for 3 complete continuous loading and unloading cycles. As illustrated in Figs.~\ref{Fig6}(a) and \ref{Fig6}(c), the model provides high-fidelity predictions of the inductance-force and inductance-pressure relationships across the operational range. Furthermore, Fig.~\ref{Fig6}(b) demonstrates that although the model does not explicitly incorporate actuator length as an independent variable, it implicitly captures inductance variations associated with actuator deformation during both loading and unloading phases. The three-dimensional visualization in Fig.~\ref{Fig6}(d) further confirms that the model generalizes well across the entire workspace, accurately representing the coupling between inductance, force, and pressure. Quantitatively, the model achieves a coefficient of determination $R^2=0.9591$ and a root-mean-square error (RMSE) of approximately 0.0294 $\mu$H. This small residual error confirms that the force-pressure-based inductance model provides a reliable foundation for real-time self-sensing and closed-loop control of the FR-PTCA.

\section{Hybrid Observer Design and State Validation}

%Significant force-length hysteresis in FR-PTCA limits the effectiveness of open-loop control strategies. Since external sensors introduce additional stiffness that compromises the actuator's inherent compliance, a self-sensing framework is required to achieve precise closed-loop control while maintaining the mechanical compliance of the actuator.

\subsection{Nonlinear State Observer Design}

Although the inductance-force-pressure mapping $\mathcal{L}(F,P)$ is established in the preceding section, real-time force inversion remains challenging in practical scenarios. The raw inductance signals acquired from the self-sensing circuitry are susceptible to electromagnetic interference and high-frequency noise. Moreover, the mapping function $\mathcal{L}(F, P)$ exhibits strong nonlinearity and non-monotonic behavior. As illustrated in Fig.~\ref{Fig6}(a), the inductance-force curve is non-monotonic and hysteretic, meaning that a single inductance value may correspond to multiple possible force states. Furthermore, the sensitivity gradient $\partial \mathcal{L} / \partial F$ approaches zero near the peak of the curve, which introduces numerical instability and vanishing gradient issues for conventional root-finding methods. Finally, static inversion of the mapping neglects the temporal continuity of actuator dynamics, which potentially leads to physically inconsistent force estimates. 

To address these challenges, we propose a hybrid EKF-optimization observer. This estimator integrates a constrained nonlinear optimization step within an EKF framework, allowing the algorithm to resolve the inversion ambiguity while exploiting dynamic continuity of the actuator system.

The objective of the observer is to dynamically reconstruct the instantaneous force $F_k$ at time step $k$ from inductance measurements while maintaining physical feasibility and trajectory smoothness. First, the raw inductance signal $\mathcal{L}_{raw}$ is filtered using a third-order Butterworth low-pass filter with a cutoff frequency of $f_c= 10~Hz$ to suppress measurement noise while preserving dynamic characteristics. We define the state vector as $\mathbf{x}_k = [F_k, \dot{F}_k]^T$. During the prediction step, the EKF propagates the \textit{a priori} state estimate $\hat{\mathbf{x}}_{k|k-1}$ based on a constant velocity model. Assuming uniform motion within a small sampling interval $\Delta t$, the prediction is expressed as:
\begin{equation}
\hat{\mathbf{x}}_{k|k-1} = \mathbf{A} \hat{\mathbf{x}}_{k-1|k-1}, \quad \mathbf{A} = \begin{bmatrix} 1 & \Delta t \\ 0 & 1 \end{bmatrix}
\end{equation}
Instead of directly preforming the standard EKF measurement update, we introduce an intermediate optimization layer to solve the nonlinear inverse problem. A pseudo-measurement of force, denoted as $F^*_k$, is obtained by minimizing a composite cost function $\mathcal{J}(F)$:
\begin{equation}
F^*_k = \operatorname*{argmin}_{F \in \Omega_F} \left( \mathcal{J}_{\text{fit}}(F) + \mathcal{J}_{\text{dyn}}(F) + \mathcal{J}_{\text{reg}}(F) \right)
\end{equation}
where $\Omega_F = [F_{\min}, F_{\max}]$ defines the physically feasible range of force values, preventing the solver from converging to mathematically valid but physically implausible solutions. The cost function comprises three terms, each addressing specific challenges in the nonlinear inversion.

\textbf{Model Fidelity Term ($\mathcal{J}_{\text{fit}}$):}
\begin{equation}
\mathcal{J}_{\text{fit}}(F) = \lambda_1 \left( \mathcal{L}(F, P_k) - \mathcal{L}_k \right)^2
\end{equation}
where $\mathcal{L}_{k}$ and $P_k$ denote the measured inductance and input pressure at time step $k$, respectively, and $\lambda_1$ is a weighting coefficient. This term enforces consistency between the predicted inductance from the model $\mathcal{L}(F, P_k)$ and the actual measurement, ensuring fidelity to the electromagnetic characteristics of the FR-PTCA.

\textbf{Continuity Constraint Term ($\mathcal{J}_{\text{dyn}}$):}
\begin{equation}
\mathcal{J}_{\text{dyn}}(F) = \lambda_2 \left( F - \hat{F}_{k|k-1} \right)^2
\end{equation}
where $\hat{F}_{k|k-1}$ is the predicted force from the EKF. This term penalizes deviations from the prior predicted states and therefore restricts the optimization search to the vicinity of the physically plausible trajectory. Crucially, this constraint resolves the ambiguity associated with the multi-valued inductance-force mapping, enabling the algorithm to distinguish between the rising and falling branches of the curve.

\textbf{Peak Stability Regularization Term ($\mathcal{J}_{\text{reg}}$):}

\begin{equation}
\mathcal{J}_{\text{reg}}(F) = \lambda_3 \left( 1 - \frac{1}{1 + \gamma \left( F - \hat{F}_{k|k-1} \right)^2 } \right)
\end{equation}
where $\gamma$ is a shape parameter controlling the width of the attraction basin. This nonlinear regularization term stabilizes the optimization near the zero-gradient region of the inductance curve, preventing abrupt jumps to spurious solutions caused by measurement noise and curve symmetry. %creates a robust attraction basin; the cost is minimized when the candidate solution aligns with the velocity-predicted position, whereas a saturation penalty is imposed if the optimization algorithm attempts to jump to a mathematically valid but physically discontinuous spurious solution caused by curve symmetry. This mechanism effectively leverages the inertial momentum of the system to traverse the zero-gradient region smoothly, thereby preventing abrupt jumps induced by high-frequency signal fluctuations at the signal peak.

The optimized force estimate $F_k^*$ is then treated as the measurement input for the EKF update step. Through this hierarchical architecture, the observer combines the global search capability of the nonlinear optimization with the probabilistic filtering properties of the EKF. Together with the stiffness model introduced in Eq.~\eqref{dyn_model}, the observer enables smooth and robust real-time estimation of both force and actuator displacement.

\subsection{Validation of Self-Sensing Estimation}

To validate the effectiveness of the proposed self-sensing model and the nonlinear hybrid observer, experiments were conducted using the platform illustrated in Fig.~\ref{Fig3}(a). These tests were systematically designed to demonstrate the comprehensive validity of the proposed approach across the full operational range of the FR-PTCA, spanning from a zero-force condition to a maximum-force state. Specifically, a linear stepper motor drives the actuator from a zero-force state to a maximum strain of 170$\%$ through continuous cyclic stretching, while the internal pressure of the FR-PTCA is varied from 0 to 0.65~MPa discretely upon the completion of each cycle. Throughout this process, the nonlinear hybrid observer continuously processes the raw inductance signals to simultaneously estimate output force and the length of the actuator in real time.

\begin{figure}[htbp]
        \centering
        \includegraphics[width=0.5\textwidth]{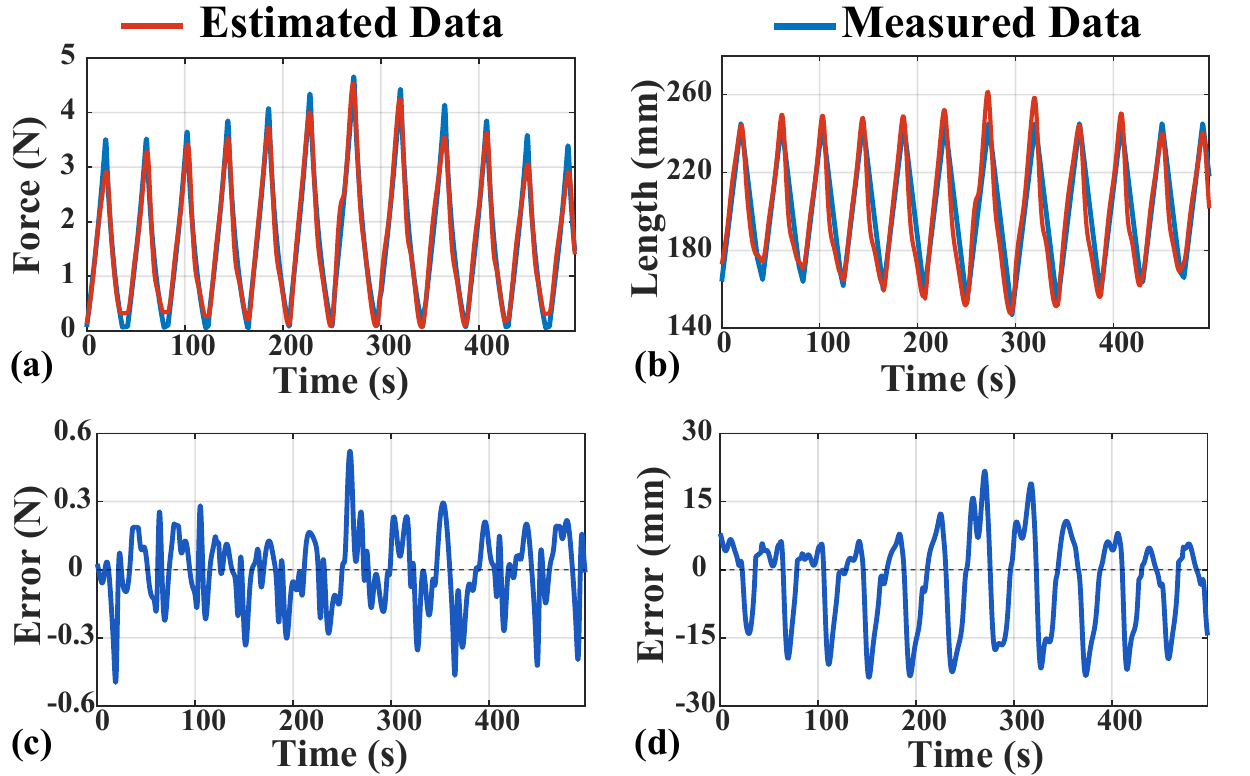}
        \caption{Experimental validation of the force and displacement estimation. (a) Measured versus estimated force over time. (b) Measured versus estimated actuator length. (c) Instantaneous force estimation error. (d) Instantaneous displacement estimation error.}
        \label{Fig7}
\end{figure}

The experimental results are illustrated in Fig.~\ref{Fig7}. Specifically, Fig.~\ref{Fig7}(a) and Fig.~\ref{Fig7}(b) compare the estimated force and length trajectories with their corresponding measured values, while Fig.~\ref{Fig7}(c) and Fig.~\ref{Fig7}(d) illustrate the corresponding estimation errors. The maximum instantaneous force estimation error remained below 0.53 N throughout the experiment. A quantitative evaluation demonstrates that the observer exhibits high accuracy in force estimation. Statistical analysis reveals an RMSE of 0.1745 N, a mean absolute error (MAE) of 0.1358 N, and a normalized RMSE (NRMSE) of 3.79$\%$. For displacement estimation, the RMSE is 10.89 mm, the MAE is 8.74 mm, and the NRMSE is 11.03$\%$.

The comparison of these metrics indicates that the force estimation accuracy is significantly higher than that of the displacement estimation. This disparity arises because the force is directly inferred from the inductance-force mapping established in Section IV, whereas the displacement is indirectly reconstructed through the dynamic model. Since the stiffness relationship in Eq.~\eqref{dyn_model} is based on a linearized approximation, it does not fully capture the complex hysteresis of the FR-PTCA. Consequently, modeling errors propagate into the displacement estimation. In addition, the largest estimation errors occur primarily near motion reversal points, as illustrated in Figs.~\ref{Fig7}(c) and \ref{Fig7}(d). These deviations are likely attributable to the inversion of the friction and strain directions during the transition between contraction and extension, which introduces transient nonlinearities that are not captured by the simplified dynamic model.

Overall, the experimental results demonstrate that the inductance-based self-sensing approach enables accurate and reliable force estimation without external sensors. Although the indirectly inferred displacement exhibits larger errors due to unmodeled hysteresis effects, the overall estimation accuracy remains sufficient for practical soft robotic control tasks. 

\section{Control Performance and Robustness}
\subsection{Self-Sensing Closed-Loop Control Experiments}

\begin{figure}[htbp]
        \centering
        \includegraphics[width=0.5\textwidth]{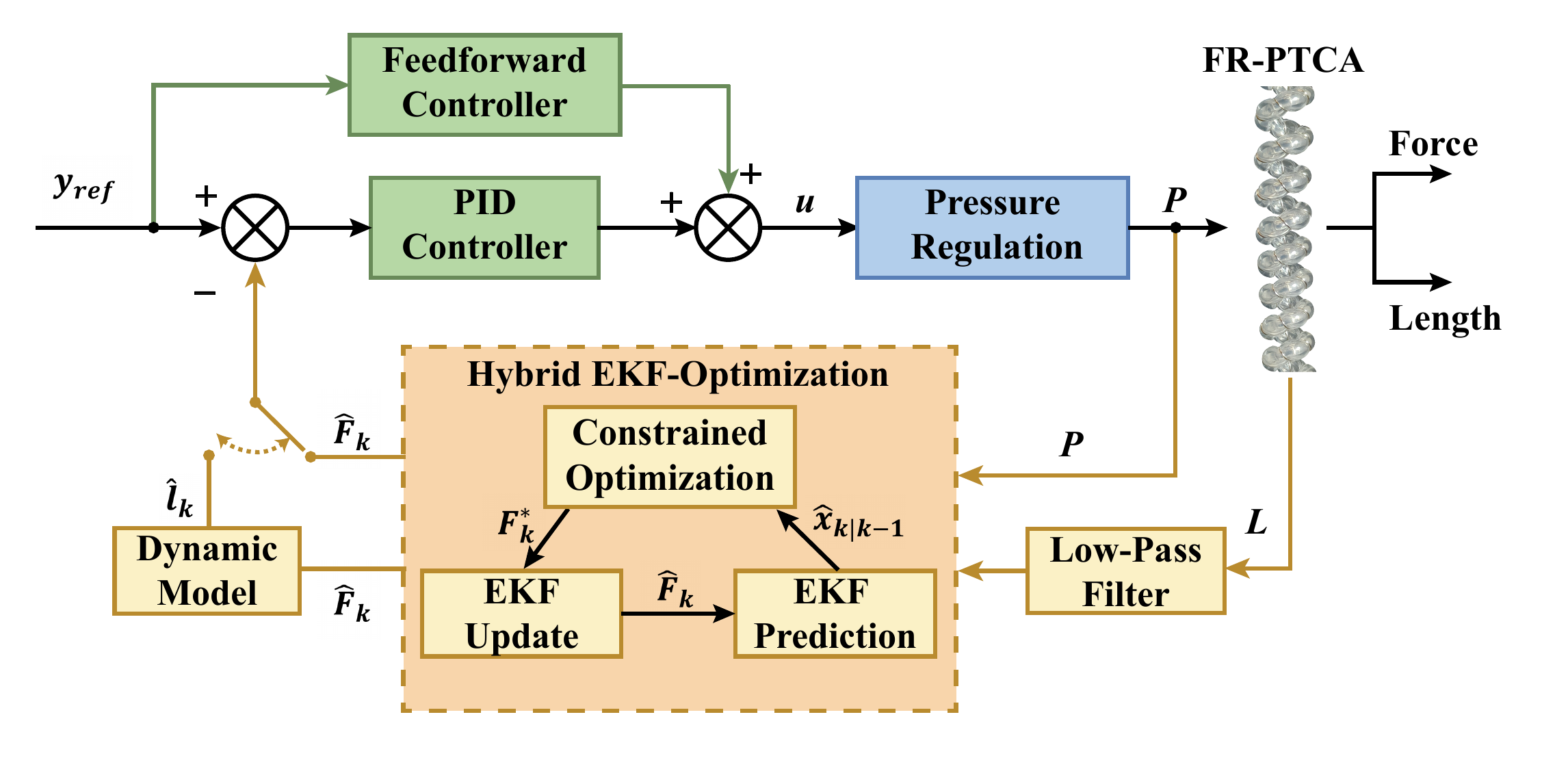}
        \caption{Block diagram of the proposed control framework.}
        \label{Fig8}
\end{figure}

To evaluate the effectiveness of the proposed self-sensing model in real-time closed-loop control, and to demonstrate the necessity of feedback for precise FR-PTCA actuation, we implemented a composite control framework that integrated proportional-integral-derivative (PID) feedback with model-based feedforward compensation, as illustrated in Fig.~\ref{Fig8}. Within this framework, the feedback loop utilizes inductance measurements to estimate actuator states through the proposed EKF-optimization observer. The control system operates at an update frequency of 20 Hz, with PID gains set to $K_P=0.027$, $K_I=0.001$, and $K_D=0.003$. Experiments were conducted under both isometric (force control) and isotonic (displacement control) conditions to compare three control strategies, including open-loop feedforward control, external sensor-based feedback control, and the proposed self-sensing feedback control.

The experimental setup for force control is illustrated in Fig. \ref{Fig3}(a). Under the isometric condition, the FR-PTCA is fixed at both ends to maintain a constant length, while the output force is regulated by adjusting internal pressure. The reference trajectories consist of sinusoidal and triangular waves with frequencies of 0.2 Hz and 0.05 Hz.

\begin{figure}[htbp]
        \centering
        \includegraphics[width=0.5\textwidth]{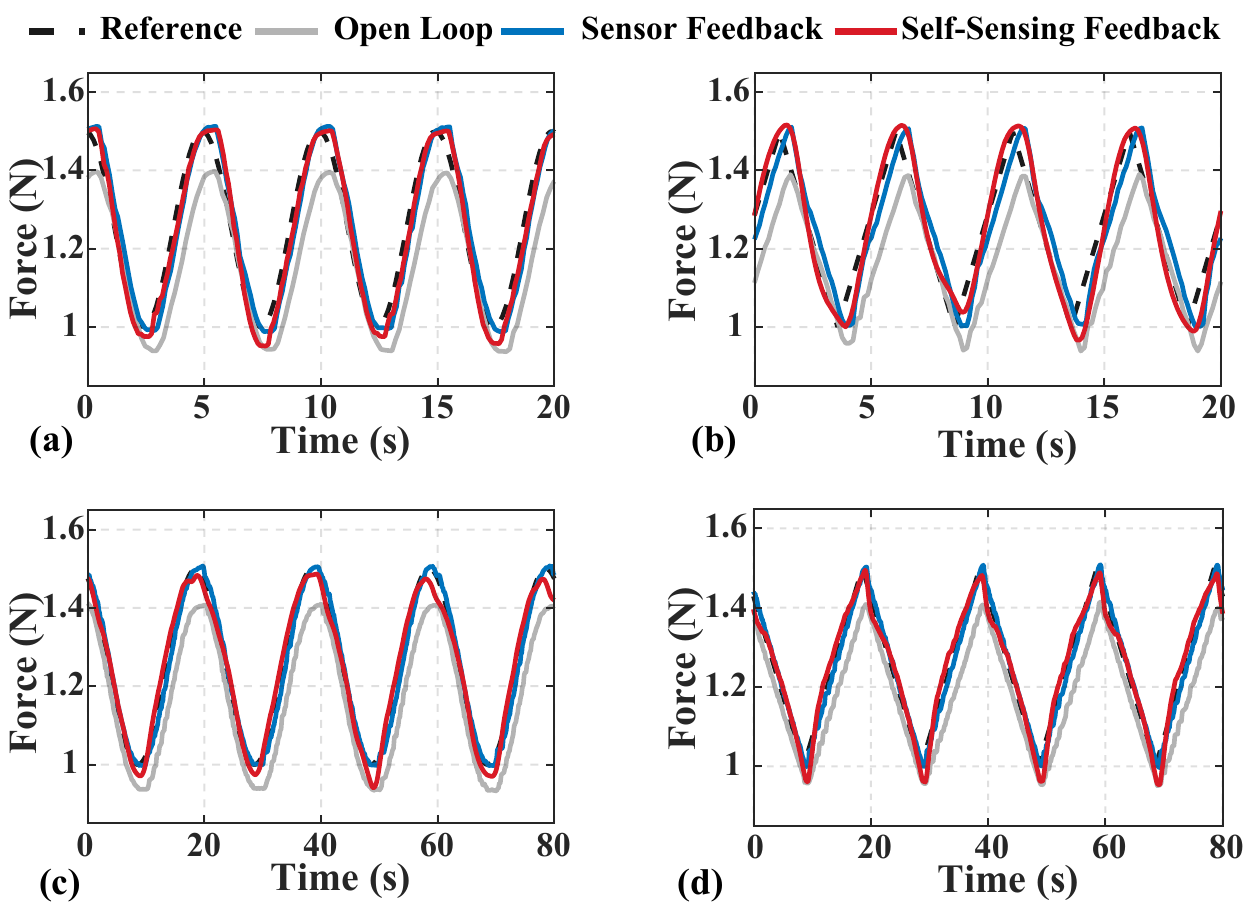}
        \caption{Experimental results of closed-loop force control. (a),(c) Sinusoidal force tracking at 0.2 Hz and 0.05 Hz, respectively. (b),(d) Triangular force tracking at 0.2 Hz and 0.05 Hz, respectively.}
        \label{Fig9}
\end{figure}

Fig.~\ref{Fig9} presents the experimental results for force tracking, with error statistics provided in Table \ref{tab:control_results}. The results indicate that open-loop feedforward control follows the trend of the reference trajectory, confirming the structural validity of the dynamic model. However, significant steady-state errors remain, since open-loop control cannot compensate for actuator hysteresis or external disturbances. The introduction of closed-loop feedback substantially improves the tracking accuracy. 

In particular, the proposed self-sensing control strategy achieves a tracking accuracy comparable to that of external sensor-based feedback control, significantly outperforming traditional open-loop control. Quantitatively, the external force sensor reduces the RMSE by 39.3\% to 79.1\% across the tested trajectories. The proposed self-sensing control achieves a comparable reduction of 42.3\% to 68.1\%, despite relying solely on inductance measurements without external instrumentation. Although the self-sensing control performs slightly below the sensor-based feedback at low frequencies, the maximum RMSE difference is limited to 0.0101 N. Conversely, at higher frequencies, the proposed method achieves lower tracking errors than the sensor-based control, demonstrating that inductance-based self-sensing provides sufficient accuracy for closed-loop force control without external sensors.

\begin{table}[htbp]
\centering
\caption{Comparison of trajectory tracking errors for different control methods}
\label{tab:control_results}
\setlength{\tabcolsep}{4pt} % Adjust column spacing
\begin{tabular}{llccc}
\hline\hline
\textbf{Trajectory} & \textbf{Method} & \textbf{RMSE} & \textbf{MAE} & \textbf{Imp. (\%)} \\ 
& & (N / mm) & (N / mm) & (RMSE) \\ \hline
\multicolumn{5}{c}{\textbf{Force Tracking (Unit: N)}} \\ \hline
\multirow{3}{*}{Sine (0.2 Hz)} 
 & Open-Loop & 0.1235 & 0.0995 & - \\
 & Sensor FB & 0.0721 & 0.0649 & 41.6\% \\
 & \textbf{Self-Sensing} & \textbf{0.0713} & \textbf{0.0624} & \textbf{42.3\%} \\ \hline
\multirow{3}{*}{Triangle (0.2 Hz)} 
 & Open-Loop & 0.1111 & 0.0899 & - \\
 & Sensor & 0.0674 & 0.0632 & 39.3\% \\
 & \textbf{Self-Sensing} & \textbf{0.0623} & \textbf{0.0470} & \textbf{43.9\%} \\ \hline
\multirow{3}{*}{Sine (0.05 Hz)} 
 & Open-Loop & 0.0860 & 0.0816 & - \\
 & Sensor & 0.0197 & 0.0178 & 77.0\% \\
 & \textbf{Self-Sensing} & \textbf{0.0275} & \textbf{0.0230} & \textbf{68.1\%} \\ \hline
\multirow{3}{*}{Triangle (0.05 Hz)} 
 & Open-Loop & 0.0814 & 0.0770 & - \\
 & Sensor & 0.0170 & 0.0162 & 79.1\% \\
 & \textbf{Self-Sensing} & \textbf{0.0271} & \textbf{0.0217} & \textbf{66.7\%} \\ \hline
\multicolumn{5}{c}{\textbf{Displacement Tracking (Unit: mm)}} \\ \hline
\multirow{3}{*}{Sine (0.05 Hz)} 
 & Open-Loop & 2.2676 & 1.9900 & - \\
 & Sensor & 0.4958 & 0.4211 & 78.1\% \\
 & \textbf{Self-Sensing} & \textbf{0.6058} & \textbf{0.4826} & \textbf{73.3\%} \\ \hline
\multirow{3}{*}{Sine (0.2 Hz)} 
 & Open-Loop & 2.5884 & 2.1886 & - \\
 & Sensor & 0.9457 & 0.7186 & 63.5\% \\
 & \textbf{Self-Sensing} & \textbf{1.2567} & \textbf{0.9787} & \textbf{51.5\%} \\ \hline\hline
\end{tabular}
\end{table}

\begin{figure}[htbp]
        \centering
        \includegraphics[width=0.5\textwidth]{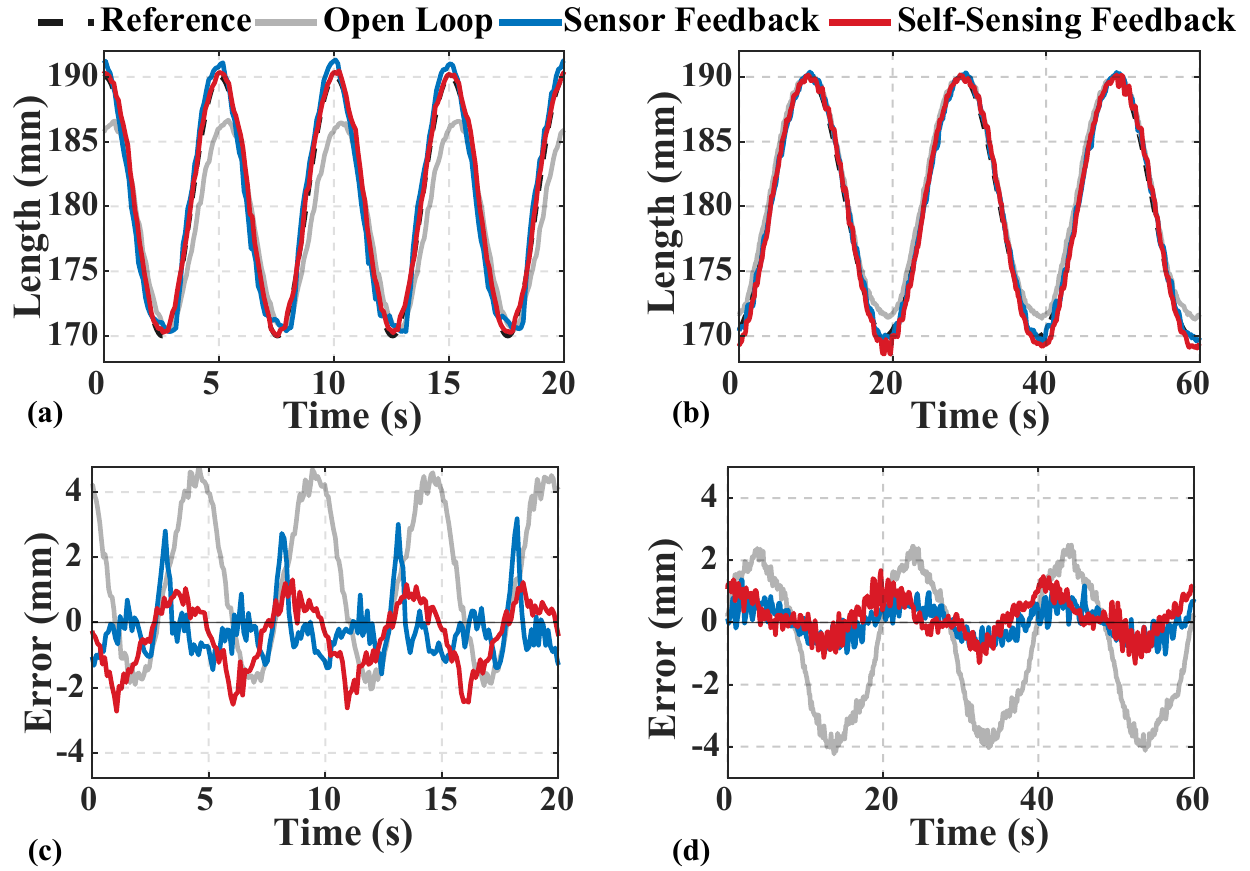}
        \caption{Experimental results of closed-loop displacement control. (a),(b) Displacement tracking for sinusoidal trajectories at 0.2 Hz and 0.05 Hz, respectively. (c),(d) Corresponding tracking errors for the 0.2 Hz and 0.05 Hz trajectories.}
        \label{Fig10}
\end{figure}

The experimental setup for displacement control is illustrated in Fig. \ref{Fig3}(b). In contrast to the isometric constraints of the force control experiments, this isotonic configuration maintains a constant external load, while the actuator length is regulated to track sinusoidal reference trajectories. A laser displacement sensor is used solely for independent performance validation. The experimental results are presented in Fig. \ref{Fig10}, with the corresponding quantitative errors detailed in Table \ref{tab:control_results}. Similar to the observations in force control, open-loop displacement tracking yields substantial deviations, with RMSE values exceeding $2$ mm. When the laser displacement sensor is incorporated into the feedback loop, the system achieves the best tracking performance, reducing the RMSE by 63.5\% to 78.1\%. In comparison, the self-sensing feedback control effectively suppresses open-loop tracking errors and reduces the RMSE by 51.5\% to 73.3\%. These results suggest that although displacement estimation is indirectly obtained through the dynamic model, the resulting accuracy remains sufficient for closed-loop motion control. The continuous state estimation provided by the hybrid observer enables reliable displacement regulation in highly compliant soft robotic systems without dedicated displacement sensors.

\subsection{External Load Perturbation Analysis}

\begin{figure}[htbp]
        \centering
        \includegraphics[width=0.5\textwidth]{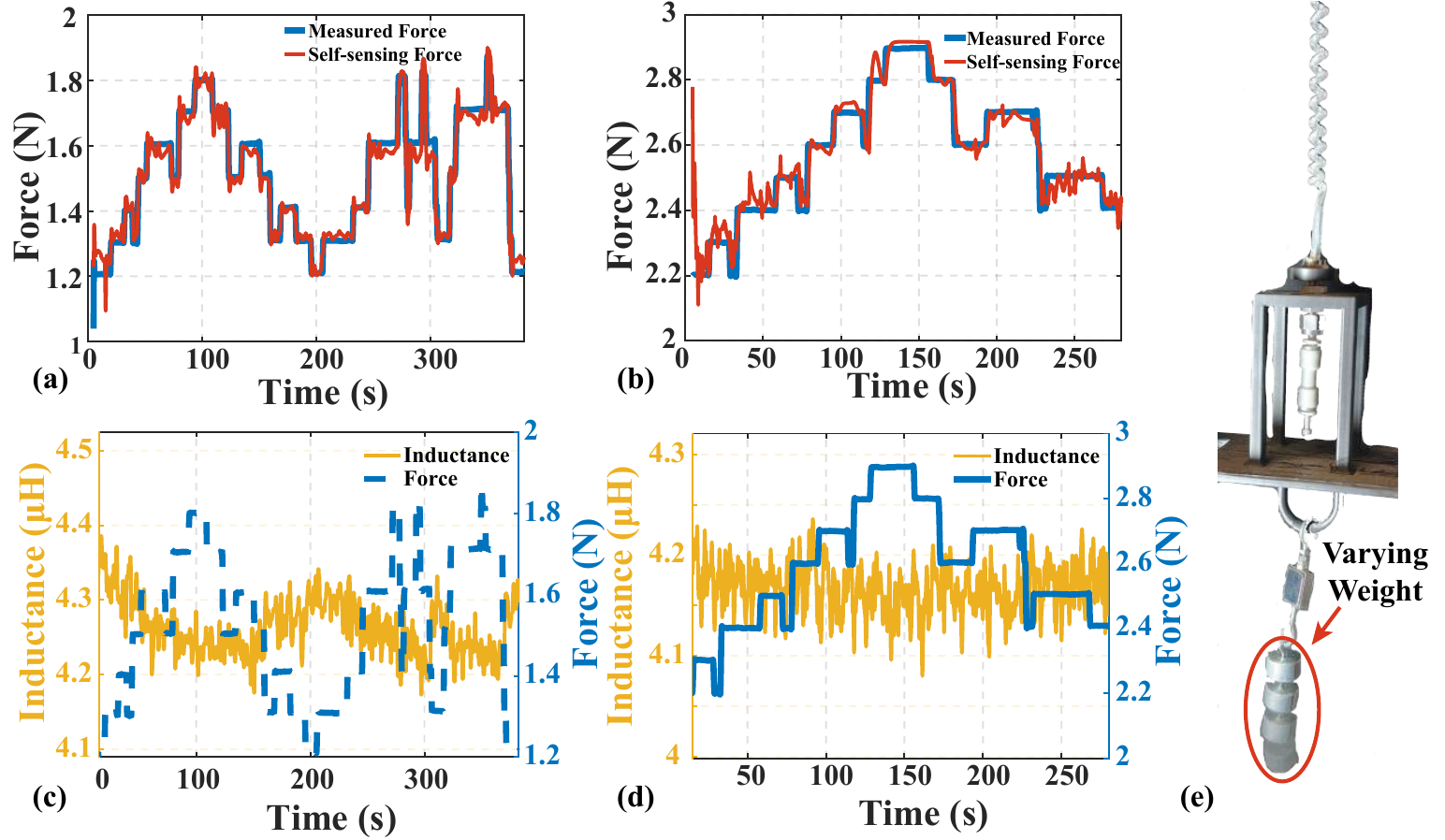}
        \caption{Experimental results of the load perturbation analysis. (a),(b) Comparison between the self-sensed force and the actual force measured by the load cell under step-change loading conditions. (c),(d) Corresponding trends trends between the measured inductance and applied force. (e) Experimental setup with randomly placed weights of varying mass.}
        \label{Fig11}
\end{figure}

To rigorously evaluate the robustness of the proposed self-sensing framework under external disturbances and its sensitivity to load variations, load perturbation experiments were conducted. The experimental setup, illustrated in Fig.~\ref{Fig11}(e), was designed to introduce controlled force disturbances. The FR-PTCA was suspended vertically, with a reference force sensor attached to the distal end. To simulate random variations in external loading, calibrated weights were randomly applied to and removed from the sensor assembly. Throughout the experiment, a closed-loop controller regulated the pneumatic pressure to maintain a constant actuator length.

The results of the load perturbation analysis are presented in Fig.~\ref{Fig11}. Figs.~\ref{Fig11}(a) and \ref{Fig11}(b) compare the self-sensed force with the ground-truth force measured by the load cell under step-change loading conditions. These results demonstrate that the inductance-based self-sensing approach rapidly and accurately captures the dynamic force variations associated with the suspended weights. Quantitative evaluation further confirms that the estimation error remains bounded within 0.09N throughout the loading and unloading cycles. The absence of significant drift indicates stable observer performance under dynamic loading conditions, achieving RMSE values of 0.026~N and 0.035~N, respectively.

The physical mechanism underlying this sensing capability is corroborated in Figs.~\ref{Fig11}(c) and \ref{Fig11}(d), which illustrate the correlated variation trends between the measured inductance and the terminal force. Notably, the inductance exhibits a deterministic shift in response to variations in the external load, even when the actuator length remains approximately constant. This observation indicates that variations in external loading induce subtle internal structural deformations within the FR-PTCA, which in turn alter the inductive characteristics of the embedded sensing element.

\section{Conclusion and Future Work}
This paper presented an integrated self-sensing FR-PTCA incorporating conductive nickel wire reinforcement. A key finding of this study is the existence of a deterministic and low-hysteresis relationship between actuator inductance and output force, in contrast to the strongly hysteretic inductance-length relationship. Leveraging this property, a parametric self-sensing model and a nonlinear hybrid observer combining constrained optimization with an EKF were developed to enable reliable real-time force estimation. Experimental results demonstrate that the proposed self-sensing approach significantly improves tracking performance over open-loop control and achieves accuracy comparable to systems using external load cells. Robust performance under dynamic load perturbations further highlights the potential of the proposed framework for sensor-minimal soft robotic systems.

Future work will focus on improving displacement estimation through rate-dependent hysteresis modeling, investigating scalability in multi-actuator systems with potential electromagnetic crosstalk, and integrating the actuator into continuum manipulators and grippers for contact-rich manipulation tasks.

%Future research will extend this framework in several key directions. First, we aim to enhance the accuracy of displacement estimation by incorporating rate-dependent hysteresis models into the sensing framework. Second, the scalability of the inductance-based sensing method will be investigated within multi-actuator arrays to address potential crosstalk issues. Finally, the proposed self-sensing FR-PTCA will be integrated into continuum manipulators or soft grippers to evaluate performance under realistic loading conditions and contact-rich interactions.

\small

\bibliographystyle{IEEEtran}
\bibliography{ref.bib}

\normalsize % 恢复正常字体大小

\newpage

\vfill

\end{document}